%% file: ms.tex
\pdfoutput=1
\documentclass[10pt]{article}
\input{hicss51-packages.tex}
\PassOptionsToPackage{hyphens}{url}
\usepackage{url}
\usepackage{breakurl}
\usepackage[breaklinks]{hyperref}

\usepackage{cleveref}
\usepackage{multirow}
\usepackage{todonotes}

\usepackage{graphicx} 
\graphicspath{ {./graphics/} }

\usepackage{xcolor}
\usepackage{adjustbox}
\usepackage{tabularx}
\usepackage[figuresright]{rotating}
\usepackage{lipsum}

\usepackage[labelfont=bf,textfont=md]{caption}

\usepackage{enumitem}

\newlist{questions}{enumerate}{2}
\setlist[questions,1]{label=RQ\arabic*.,ref=RQ\arabic*}

\usepackage[style=ieee]{biblatex}
\title{What to Prioritize? Natural Language Processing for the Development of a Modern Bug Tracking Solution in Hardware Development}

\author{Thi Thu Hang Do \\
  Karlsruhe Institute of Technology \\
  {\underline{hang.do@alumni.kit.edu}} \\\And
  Markus Dobler \\
  IBM R\&D Germany \\
  {\underline{mado@de.ibm.com} }\\\And 
  Niklas Kühl \\
  Karlsruhe Institute of Technology \\
  IBM Germany \\
  {\underline{niklas.kuehl@kit.edu}} \\}

\date{}

\begin{document}
\maketitle
\begin{abstract}
Managing large numbers of incoming bug reports and finding the most critical issues in hardware development is time consuming, but crucial in order to reduce development costs.
In this paper, we present an approach to predict the time to fix, the risk and the complexity of debugging and resolution of a bug report using different supervised machine learning algorithms, namely Random Forest, Naive Bayes, SVM, MLP and XGBoost. Further, we investigate the effect of the application of active learning and we evaluate the impact of different text representation techniques, namely TF-IDF, Word2Vec, Universal Sentence Encoder and XLNet on the model's performance. The evaluation shows that a combination of text embeddings generated through the Universal Sentence Encoder and MLP as classifier outperforms all other methods, and is well suited to predict the risk and complexity of bug tickets.
\end{abstract}

\section{Introduction} \label{introduction}

The development of computer hardware in general and processors in particular is an indispensable task in modern society, as the chip shortages in the light of the COVID-19 crisis clearly emphasized \cite{akcil2020covid}. In the case of modern high-end processors, this task is a complex process that spans many years and involves up to millions of lines of code \cite{wile_comprehensive_2005}.
During this time, the unfinished logic design of the chip will regularly contain thousands of flaws and bugs.

Hence, it is no surprise that processor verification is widely regarded as bottleneck of the hardware design process \cite{adir2004genesys}. Design verification takes about 80\% of the total product development time and has become more challenging due to the demand for higher performance chips and shorter time-to-market combined with the exponential growth in hardware size \cite{stroud20091, adir2004genesys}.

Consequently, it is essential that no critical bugs remain when the chips get manufactured, because re-fabrication of hardware can cost many hundreds of thousands of dollars \cite{wile_comprehensive_2005}.
At the same time, unnecessary delays in the development process lead to a later time-to-market, also negatively affecting revenue.

Therefore, project management needs to closely keep track of remaining known bugs: In order to focus resources on the most critical bugs, estimates are needed how critical the risk of each bug is, how complex handling it will be and how long it will take to fix each bug.

Until now, this bug triage had to be done manually, which was costly in terms of invested time and oftentimes repetitive for the affected individuals.
In this work, we propose a text-mining-based method to estimate these properties automatically, relieving project management from that task.

In order to classify the bug reports, we use supervised machine learning techniques, which need to be trained with already labeled data. However, as only experts are able to label the specific tickets, their time and budget is limited. In fact, manual labeling of all data is not feasible in practice. Hence, we implemented an active learning approach which shall enable high prediction quality using few labeled data.

In this paper, we propose methods for answering the following questions:

\begin{questions}[topsep=0pt,itemsep=-1ex,partopsep=1ex,parsep=1ex]
    \item Which text representation model performs best for technical vocabulary?
    \item How much does active learning help to boost the manual labeling process?
    \item Which targets can be predicted to what extend?
    \item Which classifier gives the most accurate label prediction for a ticket?
    \item How does prediction quality improve with more ticket entries?
\end{questions}

 Finding answers to these questions, 
 we want to illustrate the benefit of natural language processing and machine learning in the domain of bug tracking in hardware development. This is special, because the language is highly technical and existing approaches cannot be simply copied. Additionally, we evaluate the feasibility of predicting the complexity (for debugging and resolution) of a bug report, which is especially important for efficient project management. Therefore, we benchmark multiple pre-processing methods and algorithm combinations with a subsequent parameter optimization and report on the results. These can inform future researchers in their design of similar applications. Finally, we show the economic impact of a successful deployment of our solution and demonstrate how the reduction of uncertainty in project management leads to improved outcomes.

The rest of our paper is structured as follows. \Cref{relatedwork} briefly describes the background and related work and \Cref{methodology} presents our experimental design. \Cref{results} shows an overview over our results and gives an interpretation of these. Lastly, \Cref{conclusion} concludes this paper.

\section{Background and Related Work} \label{relatedwork}

In this section, we provide a description of the relevant background knowledge required for our method. First, we give a short introduction to processor verification (\Cref{verification}). Second, we elaborate on active learning and the machine learning techniques relevant to our work (\Cref{activelearning}). Third, we discuss different text representation methods, which serve as the basis for our approach (\Cref{nlp}). Fourth, an overview of similar approaches and the embedding of our work is given (\Cref{similarapproaches}).

\subsection{Fundamentals of Processor Verification} \label{verification}

The initial concept for a new or improved processor is specified in a high-level design, which describes the general architecture.
To achieve high performance while at the same time keep the area and power consumption low, logic designers then implement this high-level design in low-level hardware description languages like VHDL or Verilog.
It is the job of verification engineers to compare this implementation against the high-level specification and the design intent \cite{wile_comprehensive_2005}.
For this, they make use of different verification methods:

\emph{Functional verification} runs simulations of the design under test, stimulating the inputs with different patterns each time.
Based on the high-level specification, the expected outputs are calculated and compared to the actual outputs from the simulation.
Whenever a discrepancy is detected, verification engineers need to debug the scenario in order to trace the discrepancy back to its root cause.

\emph{Formal verification} can prove the absence of bugs for certain aspects of the design by mathematically transforming the symbolic logic extracted from the design under test.
This can for example be used to ensure that modifications introduced by a performance optimization tool do not alter the functional behavior of the design.

\emph{Structural verification} checks for the correctness of (semi-)regular connection structures.
Examples for such structures are
  the on-chip clock distribution infrastructure,
  structures for the initialization of the chip,
  and
  structures that facilitate testing for manufacturing errors to filter out defective chips.
Structural verification can detect connectivity problems significantly faster than functional verification, and it often directly points to the root cause.
It can check large and distributed structures spanning the whole chip, which would be infeasible with formal verification methods because their computational complexity grows exponentially with the state space.

One way to perform structural verification is to first traverse the net list (i.e. the connections between components and logic blocks) and convert it into tokens.
A parser then matches these tokens against a grammar that describes the expected structures.
This approach allows to encode the specification in a higher-level grammar instead of having to account for every allowed variation in the low-level implementation \cite{jaschke_static_2017}.

Whenever verification engineers expose a bug, they report it in a bug tracking system, so that the logic designers know what they still need to fix.
Compared to e.g. news texts or social media posts, discussions in bug tracking tickets contain a mix of 'natural' language, technical jargon, error dumps from log files and code snippets.
Because of these peculiarities, it is not clear from the outset how well natural language processing methods will be able to process such bug tracking tickets.

%

\subsection{Active Learning and Machine Learning Techniques} \label{activelearning}

Active learning results in a maximal increase of the model's performance using fewer labeled training data, if the model is allowed to select the data instances which it learns \cite{settles_active_2012}. 
In pool-based active learning, an initial classifier is trained based on the small set of labeled data. It is then applied to the unlabeled data set. Using the approach of uncertainty sampling, the least certain instance is passed to the expert for labeling and the learner is retrained \cite{lewis_heterogeneous_1994}. For measuring uncertainty, Shannon's entropy can be used \cite{shannon_mathematical_1948}. It is the most general uncertainty sampling strategy and determines the average information content of a variable \cite{settles_active_2012} by summing over all possible values of a random variable $X$ and taking into account their occurrence probability $p(x_{i})$ as shown in \Cref{eq:entropy}.

\begin{equation} \label{eq:entropy}
    H(X) = -\sum \limits_{i=1}^{n} p(x_{i}) \log_{}{p(x_{i})} 
\end{equation}


There are multiple supervised learning algorithms which can be used as classifiers. While Random Forest \cite{breiman_random_2001}, SVM \cite{cortes_support-vector_1995} and Naive Bayes \cite{rish2001empirical} models are able to reach a high prediction accuracy in a short time, 
Multi-Layer Perceptron (MLP) \cite{bishop_neural_1995} and XGBoost \cite{chen_xgboost_2016} are methods which are able to learn more complex problems. In order to improve a classifier's performance, hyperparameter optimization can be applied using one of multiple possible search algorithms \cite{bergstra2011algorithms}. 

\nocite{wang_empirical_2020}
\nocite{tran_analysis_2020}
\nocite{mikolov_efficient_2013}
\nocite{cer_universal_2018}
\nocite{yang_xlnet_2019}

\begin{sidewaystable}[htbp]
\centering
\resizebox{\textheight}{!}{
\small
\fontsize{9.7}{9.7}\selectfont
\begin{tabular}{@{}p{4cm}p{4cm}p{4.5cm}p{5cm}p{4.2cm}@{}}
\hline & \\[-1.1ex]
\textbf{Paper} &
  \textbf{Data Source} &
  \textbf{Pre-processing} &
  \textbf{Model} &
  \textbf{Targets} \\[1.1ex] \hline & \\[-1.1ex]
Menzies and Marcus (\citeyear{menzies_automated_2008}) &
  NASA Project &
  TF-IDF &
  Rule Learning &
  Severity \\
Lamkanfi et al. (\citeyear{lamkanfi_predicting_2010}) &
  Eclipse, Gnome, Mozilla &
  Term Availability (binary) &
  NB &
  Severity \\
Anvik, Murphy (\citeyear{anvik_reducing_2011}) &
  Eclipse &
  Term Frequency &
  \textbf{SVM}, NB, C4.5, EM, Rules, KNN &
  Appropriate developer \\
Lamkanfi et al. (\citeyear{lamkanfi_comparing_2011}) &
  Eclipse, Gnome &
  \textbf{Term Frequency}, TF-IDF &
  \textbf{NBM}, NB, KNN, SVM &
  Severity \\
Kanwal and Maqbool (\citeyear{kanwal_bug_2012}) &
  Eclipse &
  Term Frequency &
  NB, \textbf{SVM} &
  Priority \\
Sharma et al. (\citeyear{sharma_predicting_2012}) &
  Eclipse, Open Office &
  TF-IDF &
  SVM, NB, KNN, \textbf{ANN} &
  Priority \\
Somasundaram and Murphy (\citeyear{somasundaram_automatic_2012}) &
  Eclipse, Mozilla &
  TF-IDF, LDA &
  SVM &
  Appropriate developer \\
Alenezi and Banitaan (\citeyear{alenezi_bug_2013}) &
  Eclipse, Firefox &
  TF &
  NB, \textbf{RF}, DT &
  Priority \\
Alenezi et al. (\citeyear{alenezi_efficient_2013}) &
  Eclipse, NetBeans, Maemo &
  LOR, \boldmath$\chi^{2}$, TFRF, MI, DFS &
  NB &
  Appropriate developer \\
Zhang et al. (\citeyear{zhang_predicting_2013}) &
  CA Technologies &
  Term Frequency &
  Markov, Monte Carlo, KNN &
  Fixed bug count, fixing time \\
Otoom et al. (\citeyear{otoom_severity_2016}) &
  Eclipse, Mozilla &
  Normalized Term Frequency &
  \textbf{AdaBoost}, NB, RBF Networks, FT, RT, RF &
  Severity \\ 
Fan et al. (\citeyear{fan_where_2017}) &
  Github ITS &
  TF-IDF &
  SVM, RF, LR, NB &
  Bug or not \\
Alonso-Abad et al. (\citeyear{alonso-abad_label_2019}) &
  Github projects &
  Boolean, TF, logTF, IDF, TF-IDF &
  SVM, NBM &
  Tagging of unlabeled tasks \\
Kallis et al. (\citeyear{kallis_ticket_2019}) &
  Github projects &
  Term Frequency &
  fastText &
  Tagging of unlabeled tasks \\
Sahin, Tosun (\citeyear{sahin_conceptual_2019}) &
  CVE vulnerability database &
  Word2Vec &
  CNN, XGBoost, \textbf{LSTM} &
  Severity \\
Tan et al. (\citeyear{tan_bug_2020}) &
  Stack Overflow &
  word-POS, Word2Vec &
  LR, \textbf{NB}, KNN, LSTM &
  Severity \\
Tran et al. (\citeyear{tran_analysis_2020}) &
  \mbox{Bugzilla, Launchpad,} Mantis, Debian &
  TF-IDF &
  DT,\textbf{RF} &
  Severity, priority \\
This work &
  ClearQuest &
  \mbox{TF-IDF, Word2Vec, XLNet, USE} &
  RF, NB, MLP, SVM, XGBoost &
  \mbox{Fixing time, 
  risk, complexity} \\[1.1ex] \hline
\end{tabular}}
\vspace*{-3mm}
\caption{Overview of related work in the field of bug tracking data analysis}
\label{tab:sota}
\end{sidewaystable}

\subsection{Natural Language Processing} \label{nlp}

Bug tracking tickets consist of structured data (e.g. state, severity) and a stream of free-text discussion entries.
In order to transform this unstructured text into normalized and structured data, several text representation techniques are available. A survey of \citeauthor{wang_empirical_2020} (\citeyear{wang_empirical_2020}) compares several common language models on noisy text data \cite{wang_empirical_2020}. 

Term frequency-inverse document frequency (TF-IDF) is a term weighting scheme, which measures the importance of a word with respect to the documents in a collection. Therefore, the ratio between the occurrence frequency of keywords over the total number of keywords of a document (TF) and the inverse of the occurrence frequency of keywords over the total number of documents (IDF) is calculated. The product of both statistics results in TF-IDF values, representing a set of keywords which best describe the document \cite{tran_analysis_2020}.

Another technique to represent words as vectors is Word2Vec \cite{mikolov_efficient_2013}. There are two prevalent ways how to use neural network architectures to return word embeddings: While the continuous bag-of-words model (CBOW) uses the surrounding words (context) to predict the current word, the skip-gram model is trained to predict the context using the current word.

The Universal Sentence Encoder (USE) is a deep learning model which includes two pre-trained encoders: the transformer and the deep averaging network (DAN) \cite{cer_universal_2018}. This allows to include more context and hence, sentence embeddings can be generated.

XLNet is a large-scale language model, which also provides vector embeddings of sentences \cite{yang_xlnet_2019}. The pre-trained autoregressive transformer combines the advantages of autoregressive and autoencoding methods while avoiding their limitations. The model enables each position to use contextual information from all positions, since the context can consist of words from left and right.

\subsection{Similar Approaches} \label{similarapproaches}

\begin{figure*}[!htbp]
    \centering
    \includegraphics[width=\textwidth,keepaspectratio]{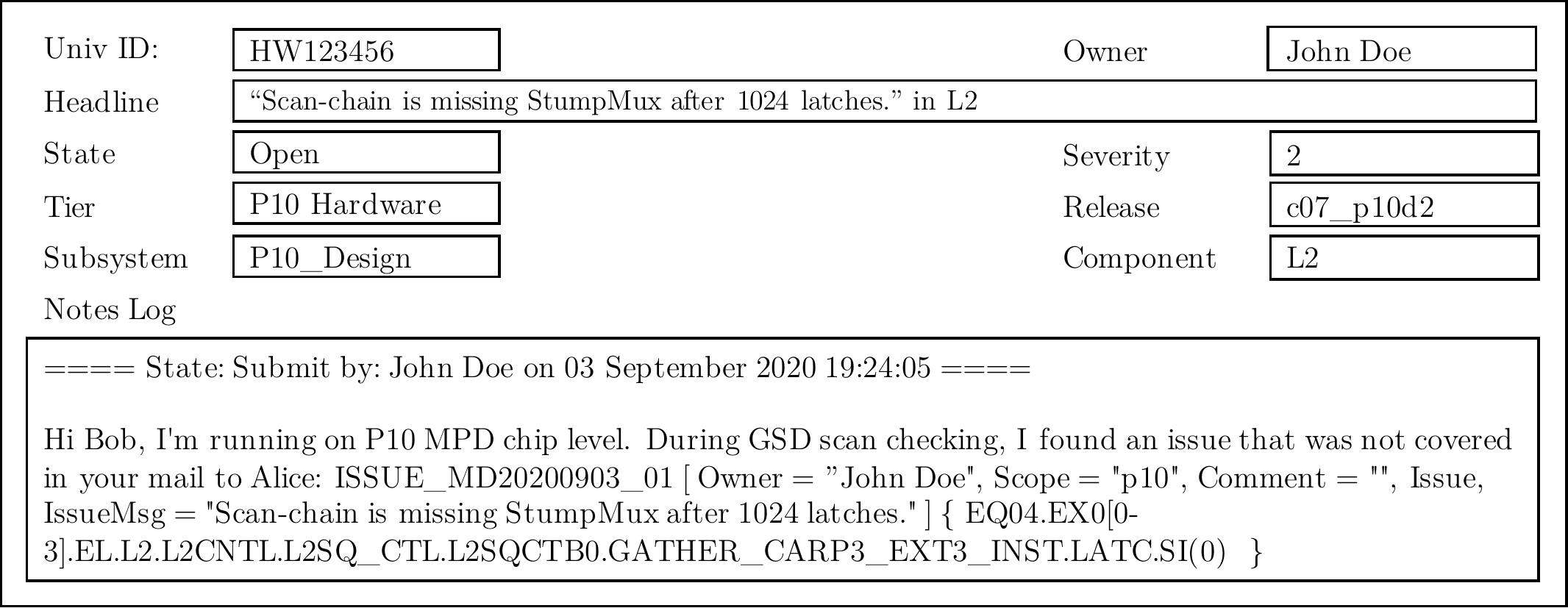}
    \caption{Example of an issue report}
    \label{fig:ticket_example}
\end{figure*}

\Cref{tab:sota} gives an overview on state-of-the-art approaches regarding the analysis of bug tracking data. Bold highlighted text shows the best performing pre-processing or classification model, if multiple options are given. Two papers try to forecast different labels with different models, so there is not just a single highest performing model and therefore there is no highlight \cite{zhang_predicting_2013,alonso-abad_label_2019}.

Most of the approaches use bug tracking data of software projects and TF-IDF as pre-processing method, or other term frequency based approaches. Only two projects decided to apply Word2Vec as language model \cite{sahin_conceptual_2019,tan_bug_2020}, but there is no related work using deep learning techniques for text representation as described in \Cref{nlp}. A great variety of machine learning models is used for solving classification tasks. Naive Bayes and SVMs are the most commonly used classifiers and perform best. The learners are generally used to predict one to a maximum of two labels, ranging from severity or priority classification, over bug fixing time prediction to recommending an appropriate developer for each bug.
    
\section{Methodology} \label{methodology}

\begin{figure*}[!t]
\centering
\includegraphics[width=\textwidth,keepaspectratio]{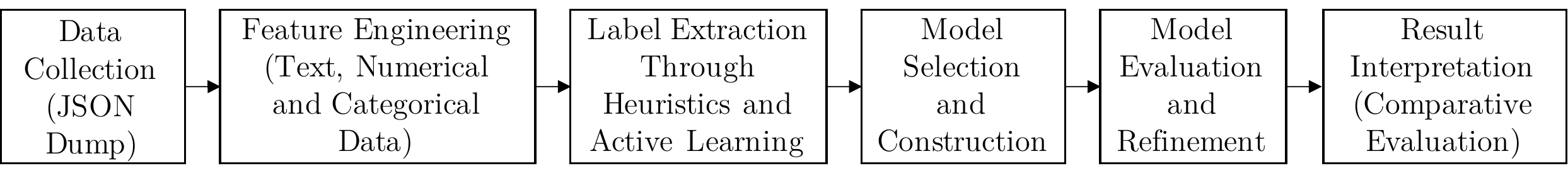}
\caption{Research design}
\label{fig:methodology}
\end{figure*}

The input data was generated during processor verification and we focused on issues discovered by the structural verification tool described in \Cref{verification}. A sample ticket can be found in \Cref{fig:ticket_example}. After extracting features from raw data, some labels can be extracted directly from the ticket data itself and some need to be manually labeled with the help of active learning (\Cref{labelextraction}). Through pre-tests we chose the best performing models, which were then used for further improvement through hyperparameter tuning. \Cref{fig:methodology} shows an overview over the mentioned steps. In the following, the steps from feature engineering to model selection are explained in more detail.

\subsection{Pre-Processing and Feature Engineering} \label{featureengineering}

\begin{figure*}[!t]
    \captionsetup{justification=centering}
    \centering
    \includegraphics[width=\textwidth,keepaspectratio]{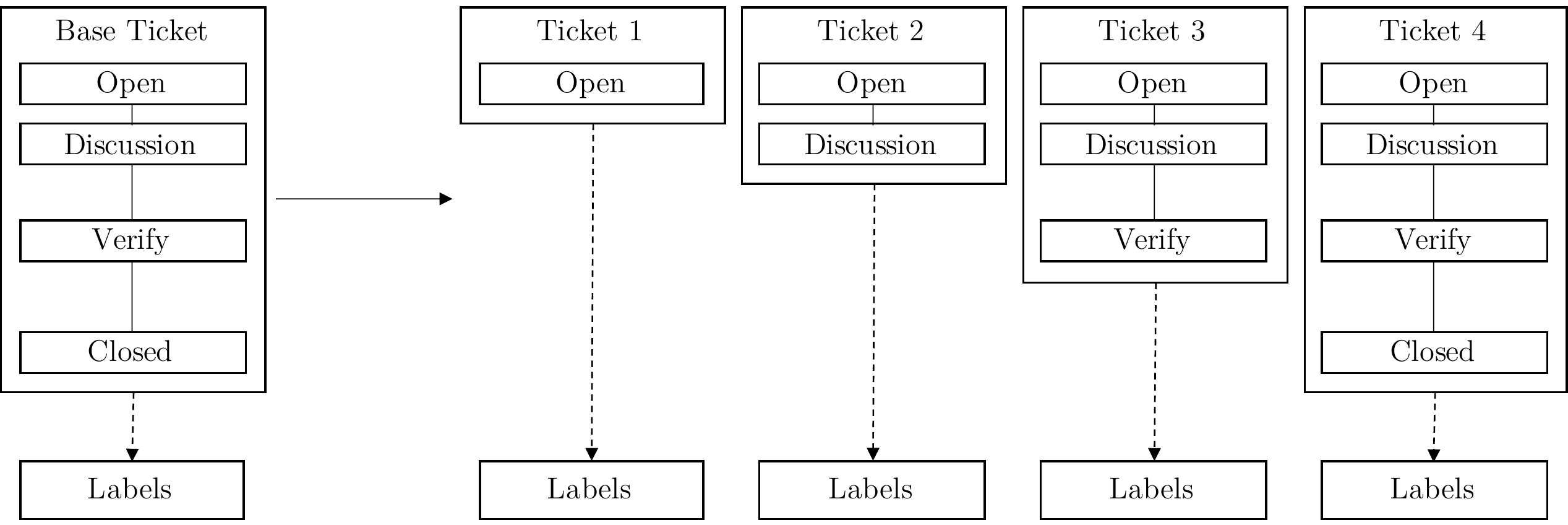}
    \caption{Process of generating multiple tickets from base ticket while avoiding data leakage (no pollution of test set with tickets that are also used in training)}
    \label{fig:ticket_generation}
\end{figure*}

We used bug tracking data from 2013 to 2020, which resulted in a total of 5,007 completed tickets. Then, the data collection was split into a train (4,506 tickets) and a test set (501 reports = 10\%) by assigning every 10th ticket id to the test set. Each bug ticket consists of different number of timestamps which document the time when a change occurred, e.g. when a comment has been added or the state of the ticket changed from `working' to `fixed'. Since the model should be able to give a prediction for tickets which are still in progress, we used the completed tickets as base tickets to generate `open` tickets: For each completed ticket, we split the ticket history and created an `open' ticket that only contained the first part of the history. As we know which base ticket belongs to the new ticket, we still know what the outcome will be in the end for the `open' ticket, and thus can generate training labels.
For example, if a base ticket has a total of 4 timestamps, 4 'artificial' tickets are created after feature engineering as illustrated in \Cref{fig:ticket_generation}: the first ticket consists of data from the first timestamp, the second ticket consists of data from the first and second timestamps, the third ticket consists of data from the first three timestamps and lastly, the fourth ticket consists of all timestamps.
From the 4,506 base tickets, we thus generated 40,661 tickets.  To avoid data leakage, we made sure that generated tickets from the same base ticket ended up either in (cross validation) training or test sets, but never in both.

The raw data comprises 128 categorical, numerical, text and date fields. After removing fields with a high proportion of NaN values ($>$ 90\%) and based on domain knowledge, 49 relevant fields are left for the feature extraction process.
For categorical data (e.g. state, action), one-hot encoding was used if there were less than 10 classes. If there were more than 10 classes, an expert created a mapping, in order to reduce the size of possible characteristics. Important time information (e.g. timestamp of change, discussion update) was used to calculate the time span between the dates. Those time periods were then used to get the minimum, maximum and mean. For numerical data (e.g. number of field changes, seconds since last state change) no further operations were needed. 
Based on the empirical study of \citeauthor{wang_empirical_2020} (\citeyear{wang_empirical_2020}), we chose the best performing language models XLNet and USE in order to process our text data (e.g. discussion entries, headlines). The study used twitter data with noisy and short messages \cite{wang_empirical_2020}, i.e. characteristics that our discussion entries also exhibit.
For XLNet and USE, the only text pre-processing step was the removal of HTML tags. Moreover, we included TF-IDF and Word2Vec, since those techniques were commonly used in state-of-the-art approaches (\Cref{tab:sota}).
For these, the text pre-processing includes numbers, punctuation and stop word removal and tokenization. To reduce the dimension of the feature vector, we chose the 200 words with the highest TF-IDF values. We averaged the output vectors resulting from Word2Vec over all words in the ticket, aiming to generate equal length vectors.

\subsection{Label Extraction} \label{labelextraction}

There are four different targets which need to be predicted: Risk, complexity of debug, complexity of resolution and time to fix. 
The fixing time which denotes the days passed until a ticket is closed, can be extracted from raw data. After calculating the fixing time for each timestamp, we assigned classes using the quantiles [0.2, 0.4, 0.6, 0.8], which correspond to the days [1, 10, 27, 65].

The remaining targets need to be labeled through experts with the help of active learning.
It is used to facilitate the manual labeling process. The risk target, which uses the final resolution as a proxy for how severe the chip design would be affected if the bug would not be fixed, can be divided into 6 classes: hardware fix, code fix, setup fix, waiver, user error, duplicate. The most critical bugs need hardware fixes, while the least critical bug reports are duplicates of existing bug tickets. A waiver is a reported `bug' that is invalid (e.g. if a general design rule does not apply for a certain block of logic) or accepted (because it technically violates a design rule but does not affect the performance and functionality of the processor). Further the complexity of debugging and of resolution shall be predicted on a scale from 0 (low) to 10 (high).


The input of active learning corresponds to the base tickets, which include the whole history of the bug ticket, since the label is the same at each observed time of a ticket's lifetime. Therefore, no multiplication of the input data is achieved as described in \Cref{featureengineering}.
To select the model used in the active learning phase, an initial pool of 39 labeled tickets was provided. TF-IDF was chosen as text representation technique, because the values can be generated fast for new data. 
Since we needed to quickly re-train the model in each iteration of the active learning loop, we selected Random Forest, Naive Bayes and SVM as potential candidates.
Their performances were compared to each other using leave-one-out cross validation because of the low number of labeled tickets. For measuring the uncertainty over a target we chose the entropy, because its objective is to minimize the log-loss. In other words, instances for which only one of the labels is highly unlikely are not favored \cite{settles_active_2012}. When active learning was implemented, we trained the selected model on a pool of 230 labeled tickets which were labeled randomly after the model selection process. For each target an own model was trained and the entropy was calculated. Afterwards, we proposed the ticket with the overall maximum entropy for one of the three targets to the expert. 

\begin{table}[!b]
\begin{tabular}{@{}ll@{}}
\hline
\textbf{Parameter} & \textbf{Search Space}          \\ \hline
max\_depth         & {[}10, 20, 30, 40, 50, None{]} \\
max\_features      & {[}auto, sqrt, log2{]}   \\
n\_estimators      & {[}10...1000{]}                  \\
criterion          & {[}gini, entropy{]}        \\ \hline
\end{tabular}
\caption{Search space for Random Forest tuning}
\label{tab:searchrf}
\end{table}
\begin{table}[!b]
\resizebox{\columnwidth}{!}{
\begin{tabular}{@{}ll@{}}
\hline
\textbf{Parameter}   & \textbf{Search Space}               \\ \hline
hidden\_layer\_sizes & {[}10...300{]}                      \\
alpha                & {[}exp(-8*log(10))...exp(3*log(10)){]} \\
activation           & {[}relu, logistic, tanh{]}    \\
solver               & {[}lbfgs, sgd, adam{]}        \\ \hline
\end{tabular}}
\caption{Search space for MLP tuning}
\label{tab:searchmlp}
\end{table}    
\subsection{Models} \label{models}

While the objective for choosing a model for active learning was to reduce run-time, our final model was selected based on optimizing the prediction quality using a 5-fold cross validation. Therefore, the ticket ids were randomly shuffled and then split into five folds of almost equal size, in order to prevent that the same base ticket is present in the training and test set. We trained each of the models (Random Forest, Naive Bayes, SVM, MLP and XGBoost) using each of the different text representation input vectors from \Cref{featureengineering}, plus a "no NLP" input vector that only contained numerical and categorical features. Afterwards we tuned the hyperparameters of the best combinations using the search spaces mentioned in \Cref{tab:searchrf} for the Random Forest and \Cref{tab:searchmlp} for our MLP with the help of the TPE algorithm \cite{bergstra2011algorithms}.

\subsection{Evaluation} \label{evaluation}

We evaluate our classifier's performance by using the widely used f1-score as follows:

\begin{equation}
f_1\textrm{-score} = \frac{\textrm{TP}}{\textrm{TP} + \frac{1}{2} (\textrm{FP}+\textrm{FN})}
\end{equation}

We compare our models with the random guesser (all classes have the same probability of being chosen) as baseline. We tried different baseline approaches beforehand and chose the random guesser, since it yields the highest f1-score among all baseline approaches on our training and test data.

\section{Results and Discussion} \label{results}

In this section we answer the research questions introduced in \Cref{introduction} and give an interpretation of the results.

\begin{figure*}[!t]
    \centering
    \includegraphics[keepaspectratio, scale=0.65]{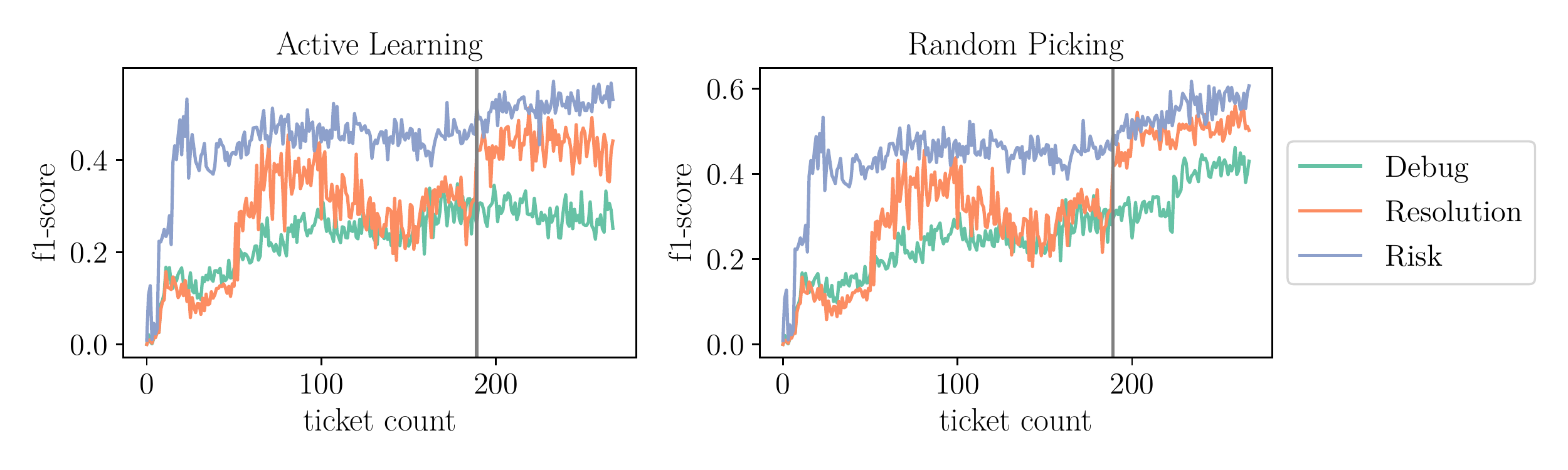}
    \vspace*{-8mm}
    \caption{Change of f1-score when labeling through active learning or random picking}
    \label{fig:activeevaluation}
\end{figure*}

\begin{table}[!b]
\resizebox{\columnwidth}{!}{
\begin{tabular}{@{}lccc@{}}
\hline
 & \multicolumn{1}{l}{\textbf{Baseline}} & \multicolumn{1}{l}{\textbf{TF-IDF \& RF}} & \multicolumn{1}{l}{\textbf{USE \& MLP}} \\ \hline
Time to Fix           & 0.20          & 0.28          & 0.29          \\
Risk                  & 0.19          & 0.40          & 0.66          \\
Debug                 & 0.04          & 0.36          & 0.60          \\
Resolution            & 0.04          & 0.36          & 0.51          \\ \hline
\textbf{Average}      & \textbf{0.09} & \textbf{0.35} & \textbf{0.51}
\end{tabular}}
\captionsetup{justification=centering}
\caption{Performance comparison of the best performing classification techniques on the hold-out set}
\label{tab:testresults}
\end{table}

\subsection{RQ1: Evaluation of NLP Techniques}

Our pre-tests showed that the performance of our text mining techniques depends on the target and the classifier used. Comparing all methods, XLNet and "no NLP" consistently exhibited the worst scores, followed by Word2Vec. Regarding the prediction quality on our training set the combination of TF-IDF and Random Forest outperforms USE and MLP when predicting the fixing time, while the latter is better regarding the prediction of the targets labeled through active learning. Therefore we chose to optimize both approaches, which led to an overall better performance of the method using USE embeddings over all targets as shown in \Cref{tab:testresults}.

Since we trained Word2Vec on our training set (total of 4,506 tickets), the poor result can be explained by the fact that the model did not have enough data available to capture relationships between the technical words. Even though we used the pre-trained XLNet and USE models and both methods are regarded as state-of-the-art approaches in the field of NLP, they performed very differently:
USE turned out to be one of the best models while XLNet is on a par with not using any NLP at all, i.e. it did not increase prediction performance beyond what is already achievable with only categorical and numerical data. A possible reason might be that XLNet was trained on text extracted from book and movies \cite{yang_xlnet_2019}, while USE used human-written English sentences extracted from a variety of web sources (e.g. discussion forum, question and answer pages) \cite{cer_universal_2018}, which contained formats closer to our heterogeneous text data. The TF-IDF approach is not as good as USE, because we did not include all TF-IDF values for all terms in our feature vector (curse of dimensionality). Therefore, there is a chance that we missed important terms, while the text embeddings from USE represent the whole text input. Additionally, TF-IDF works on individual words and thus cannot incorporate context from whole sentences.


\subsection{RQ2: Evaluation of Active Learning Approach}

Our experiments revealed that the Random Forest (f1-score: $47\%$) had the best average score over all targets generated through active learning (Naive Bayes: $19\%$, SVM: $27\%$). In order to evaluate the impact of active learning on the labeling process, we simulated the active learning process and calculated the f1-score change for each ticket which is added to the pool of labeled data. This pool is then used to retrain our Random Forest. As shown in \Cref{fig:activeevaluation} no impact of active learning on the performance can be identified. In contrast, it can be seen, that active learning is performing worse with more labeled data than our baseline, since the f1-score stays on the same level, while it increases with the number of labeled data generated through random picking.

The reason is  that the choice of entropy as uncertainty measure is not appropriate for our use case, since this method often proposes outliers (e.g. a ticket with parts of the discussion in another language; or a ticket that contains discussion threads for three independent bugs). These outliers are very dissimilar to any other ticket and thus lead to a less generalizable training set.

\subsection{RQ3: Prediction Performance for the Targets}

In this study, we investigated how accurate the targets risk, complexity and time to fix can be predicted. It can be stated that our models have the highest f1-score predicting the risk of a bug report (\Cref{tab:testresults}). 
Regarding the complexity targets, the debug complexity can be better predicted than resolution complexity. Both of our models predict the time to fix the worst.

A possible reason for the bad performance for the fixing time prediction might be that there are too many classes which can be chosen in order to meet the business requirements. Our experiments showed that we can yield a better performance level if our target has less classes. Another reason might be that our models are not complex enough to capture the underlying relationship between our features and our fixing time target. Moreover, there are external factors which can influence the fixing time which are not represented by our feature vector. For example, the fixing time also depends on the overall workload of the team, which is not reflected in the individual tickets.
The predictions for complexity are not as good as the prediction of risk, since a ticket's label was chosen based on the expert's subjective placement on a scale, so there is a possibility that users might assign slightly different values. For the risk target on the other hand, we have dedicated classes where the experts will always choose the same label, and thus our model was able to predict them well.

\subsection{RQ4: Classifier Selection}

\begin{figure*}[!htbp]
    \centering
    \includegraphics[scale=0.5,keepaspectratio]{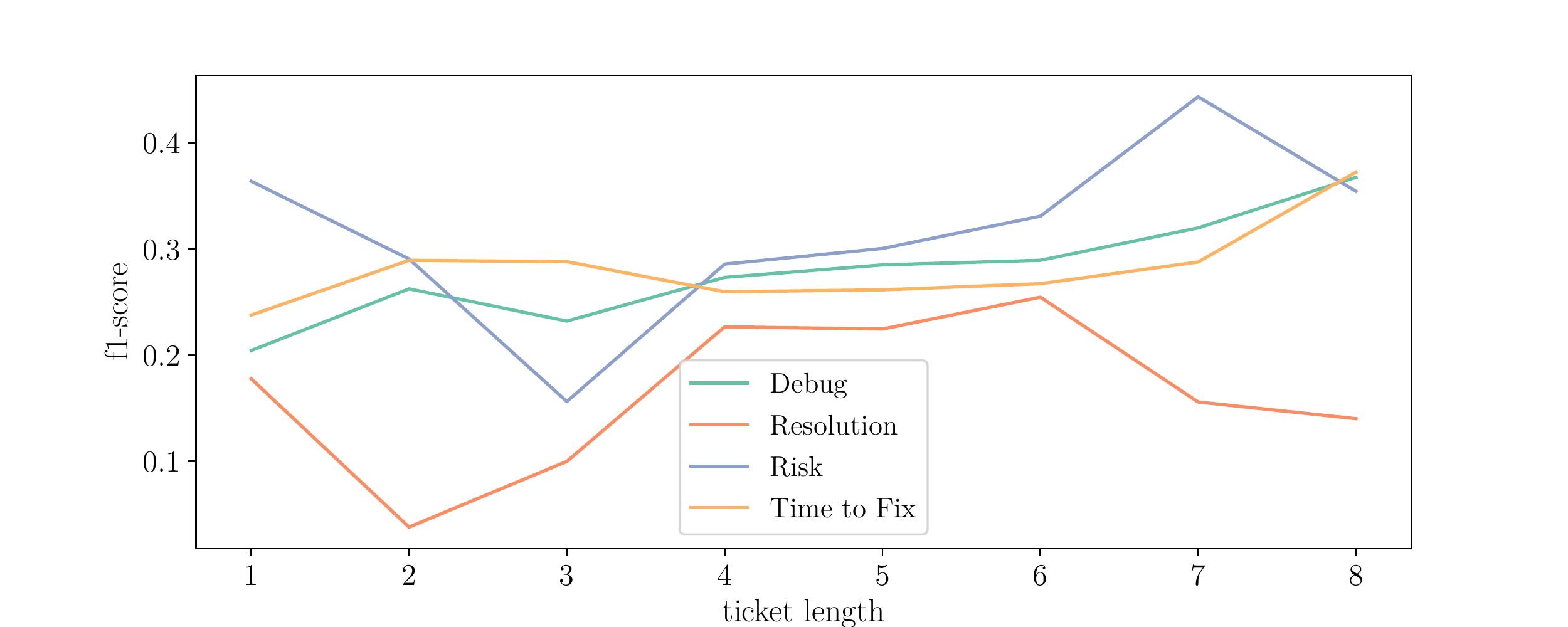}
    \caption{Development of model performance with increasing number of entries}
    \label{fig:ticket_length}
\end{figure*}

As mentioned above, the combination of TF-IDF and Random Forest and USE and MLP were outperforming our other possibilities during our pre-tests. That sets our work apart from related work, since SVM and Naive Bayes were mostly used. We chose the random guesser (all classes have the same probability of being chosen) as baseline approach. On average, the tuned models with the corresponding parameters (shown in \Cref{tab:tuningresultsrf} and \Cref{tab:tuningresultsmlp}) achieve a 4-fold improvement for Random Forest and a 6-fold improvement for MLP over the baseline approach (\Cref{tab:testresults}). 
Regarding our best performing approach (USE and MLP), the evaluation results show that we can achieve an average weighted f1-score of $29\%$ (1.5-fold improvement) for predicting the time to fix, $66\%$ (3.5-fold improvement) for risk, $60\%$ (15-fold improvement) for debug complexity and $51\%$ (13-fold improvement) for resolution complexity. 

As stated in the previous research questions, the combination of USE embeddings and MLP as classifier achieve the best performance, since it can learn more complex relationships between features and targets.


\subsection{RQ5: Prediction Quality and Ticket Length}

We chose tickets with 8 entries (which is the average ticket length) to analyze the change in prediction performance over the lifetime of the ticket. For the analysis we chose the combination of USE and MLP with tuned hyperparameters corresponding to the targets. The result is visualized in \Cref{fig:ticket_length}. For the targets time to fix, risk and resolution complexity no clear upwards trend can be identified. The trajectory of the debug complexity shows an increasing trend starting at a f1-score of $20\%$ and ending at $37\%$. 
\begin{table}[!b]
\resizebox{\columnwidth}{!}{
\begin{tabular}{@{}lcccc@{}}
\hline
 &
  \multicolumn{1}{l}{\textbf{max\_depth}} &
  \multicolumn{1}{l}{\textbf{max\_features}} &
  \multicolumn{1}{l}{\textbf{n\_estimators}} &
  \multicolumn{1}{l}{\textbf{criterion}} \\ \hline
Time to Fix & 40 & sqrt & 877 & gini    \\
Risk        & 20 & auto & 166 & gini    \\
Debug       & 10 & sqrt & 297 & entropy \\
Resolution  & 10 & sqrt & 215 & entropy \\ \hline
\end{tabular}}
\captionsetup{justification=centering}
\caption{Results of Random Forest hyperparameter tuning dependent from target}
\label{tab:tuningresultsrf}
\end{table}

\begin{table}[!b]
\resizebox{\columnwidth}{!}{
\begin{tabular}{@{}lcccc@{}}
\hline
\textbf{} & \multicolumn{1}{l}{\textbf{hidden\_layer\_sizes}} & \textbf{alpha} & \multicolumn{1}{l}{\textbf{activation}} & \multicolumn{1}{l}{\textbf{solver}} \\ \hline
Time to Fix & 36 & 3.8183 & relu & adam  \\
Risk        & 0  & 0.0332 & relu & adam  \\
Debug       & 27 & 0.1409 & relu & sgd   \\
Resolution  & 32 & 0.0005 & relu & lbfgs \\ \hline
\end{tabular}}
\captionsetup{justification=centering}
\caption{Results of MLP hyperparameter tuning dependent from target }
\label{tab:tuningresultsmlp}
\end{table}
For the risk target, the score is almost as high at the beginning as at the end of a bug lifetime, which shows that we are already able to give suitable prediction when a ticket has just been opened. This can be explained by the fact that our model can learn from log files describing the bug found which is usually attached in the first discussion entry. The slight upward trend of debugging complexity shows that with an extended number of discussion entries, it becomes more likely that the debugging complexity is higher.
On the other hand, such an effect does not show up for the resolution complexity. The resolution complexity is rather dependent on textual features than on the number of entries.
The trajectory describing the performance of the fixing time is constant except for the last entry of a ticket, when the status has been set to closed, so the model has an indicator that the remaining time is low.

\section{Conclusion} \label{conclusion}
Bug triaging is critical for the development of processors since it can help to reduce design costs. Therefore, we developed a method to ease the process of tracking the reports.
Based on a survey about text mining techniques and the experiments of state-of-the-art approaches regarding software bugs we propose USE for pre-processing and MLP as a classifier. 
Our experiments revealed that it is possible to apply the pre-trained USE model to our technical vocabulary, but the commonly used TF-IDF score also yields good results. In addition, we applied an extensive pre-testing, experimenting with the combinations of multiple pre-processing methods and algorithms followed by parameter optimization for the application in future related work. We showed that our model is able to predict the new targets regarding the complexity (debug, resolution) which cannot be found in our related work approaches. We were able to achieve a high performance level for our targets which needed to be manually labeled, even though the application of active learning did not increase our learning performance significantly.
After the successful deployment of our approach, we will be able to identify critical tickets more easily and hence, the workload of repetitive tasks will be reduced. Therefore, there will be more time for verification engineers to dedicate themselves to creative tasks. 
A limitation to our approach is the data set used for our experiments since the quality of raw data is heterogeneous. Further, we do not account for concept drift \cite{baier2019cope} in the selection of our training, validation, and test sets.
Since our active learning approach did not improve the learning rate, we propose to use other semi-supervised learning methods, or active learning with a different uncertainty measure in order to receive fewer outliers. One option is to select the ticket which most reduces the uncertainty of all the other tickets left. Finally, our method needs to be successfully deployed and the effectiveness of the predictions for project management needs to be evaluated. A promising field of research lies ahead.

\printbibliography
\end{document}

%% file: hicss51-packages.tex
\usepackage[letterpaper]{geometry}
\usepackage{hicss51}
\usepackage{times}
\usepackage[none]{hyphenat}
\usepackage{url}
\usepackage{latexsym}
\usepackage{indentfirst}
\usepackage{graphicx}
\graphicspath{{images/}}